\documentclass[10pt, conference, compsocconf]{IEEEtran}
\usepackage{cite}
\usepackage{graphicx}
\usepackage{float}
\usepackage{amsmath}
\usepackage{listings}
\usepackage{xcolor}
\usepackage{subfig}
\usepackage{url}
\usepackage{commath}
\usepackage{textcomp}
\usepackage{multirow}

\title{Deep Learning Framework to Detect Face Masks from Video Footage}

\makeatletter
\newcommand{\linebreakand}{
  \end{@IEEEauthorhalign}
  \hfill\mbox{}\par
  \mbox{}\hfill\begin{@IEEEauthorhalign}
}
\makeatother

\author{
\IEEEauthorblockN{Aniruddha Srinivas Joshi\IEEEauthorrefmark{1}, Shreyas Srinivas Joshi\IEEEauthorrefmark{2},}
\IEEEauthorblockN{Goutham Kanahasabai\IEEEauthorrefmark{3}, Rudraksh Kapil\IEEEauthorrefmark{4}, Savyasachi Gupta\IEEEauthorrefmark{5}}
\IEEEauthorblockA{B.Tech., Department of Computer Science and Engineering\\
National Institute of Technology, Warangal, Telangana, India - 506004\\
aniruddha980@gmail.com\IEEEauthorrefmark{1}, sj841871@student.nitw.ac.in\IEEEauthorrefmark{2},\\
gauthamkanags@gmail.com\IEEEauthorrefmark{3}, rkapil@student.nitw.ac.in\IEEEauthorrefmark{4}, gsavya10@gmail.com\IEEEauthorrefmark{5}}
}

\begin{document}
\maketitle

\begin{abstract}
The use of facial masks in public spaces has become a social obligation since the wake of the COVID-19 global pandemic and the identification of facial masks can be imperative to ensure public safety. Detection of facial masks in video footages is a challenging task primarily due to the fact that the masks themselves behave as occlusions to face detection algorithms due to the absence of facial landmarks in the masked regions. In this work, we propose an approach for detecting facial masks in videos using deep learning. The proposed framework capitalizes on the MTCNN face detection model to identify the faces and their corresponding facial landmarks present in the video frame. These facial images and cues are then processed by a neoteric classifier that utilises the MobileNetV2 architecture as an object detector for identifying masked regions. The proposed framework was tested on a dataset which is a collection of videos capturing the movement of people in public spaces while complying with COVID-19 safety protocols. The proposed methodology demonstrated its effectiveness in detecting facial masks by achieving high precision, recall, and accuracy.
 \end{abstract}

\section{Introduction}
\label{intro}
With the ever swift development of machine learning algorithms and methodologies in recent times, the task of face detection has been addressed to a large extent. For instance, the face detection model proposed in \cite{mref1} achieves a precision of 93\% even when detecting multiple faces. Due to the advancement of facial detectors, numerous applications such as real-time face recognition systems \cite{mref2}, security surveillance systems \cite{mref3}, etc. have been developed.\par

Despite the success of such existing techniques, there is an increasing demand for the development of robust and more efficient face detection models. In particular, the detection of masked faces proves to be a challenging and arduous task for existing face detection models due to several reasons. Firstly, traditional face detection algorithms are based on the extraction of handcrafted features. The Viola Jones face detector \cite{mref6} uses Haar features with the integral images technique to extract facial features. Other feature extraction techniques include the utilisation of the Histogram of Gradients (HOG) \cite{mref8}, Fast Fourier Transform (FFT) and Local Binary Patterns (LBP) \cite{mref7}. With advancements in the field of deep learning, neural networks can now learn features without utilising prior knowledge for forming feature extractors such as the You Only Look Once (YOLO) algorithm \cite{mref9}.\par

The pressing concern with the aforementioned approaches when it comes to face mask detection is that the face masks, with their visual diversity and various orientations behave as occlusions and variable noise to the models. This leads to a lack of local facial features, resulting in the failure of even state-of-the-art face detection models. Moreover, there is a lack of large datasets with labeled images of faces with facial masks required in order to analyse the vital characteristics common to masked faces, thus accounting for the low accuracy of existing models. These factors together justify the challenging nature of masked face detection in the field of image processing.\par

During the COVID-19 pandemic, everyone is advised to wear face masks in public \cite{mref4}. According to the World Health Organization (WHO), masks can be used for source control (worn by an infected individual to inhibit further transmission) or for the protection of healthy people. At the time of writing, the global pandemic has infected over 11 million people worldwide and has led to over half a million casualties \cite{mref5}. The wide-scale usage of face masks poses a challenge on public face detection based security systems such as those present in airports, which are unable to detect facial masks. Since the improper removal of masks can lead to contracting the virus, it has become essential to improve facial detectors that rely on facial cues, so that detection can be performed accurately even with inadequately exposed faces.\par

\section{Related works and Literature}
In this section, we review some similar works done in this domain. As elucidated in section \ref{intro}, although research on face detection has been going on for decades and has achieved great success, algorithms and methodologies that are earmarked for face mask detection are limited. \par
Ge \emph{et al.} \cite{mref10} developed a deep learning methodology to detect masked faces using LLE-CNNs, which outperforms state-of-the-art detectors by at least 15\%. In the given work, the authors introduced a new dataset called MAsked FAces (MAFA), containing 35,806 images of masked faces having different orientations and occlusion degrees. The proposed LLE-CNNs consist of three modules - proposal module, embedding module and verification module. The proposal module first combines two CNNs to extract candidate facial regions from the input image and represents them with high dimensional descriptors. After that, the embedding module is turns these descriptors into similarity based descriptors using Locally Linear Embedding algorithms and dictionaries trained on a set of faces, comprised of masked and unmasked images. Finally, the verification module is used to identify candidate facial regions and refine their positions with the help of classification and regression tasks.  \par 
Nair \emph{et al.} \cite{mref11} utilised the Viola Jones object detection framework to detect masked faces in surveillance videos. The authors argued that detecting cosmetic components such as face masks takes a significantly longer period than face detection. The framework uses the Viola Jones face detection algorithm to detect the eyes and face of subjects. If eyes are recognised and later the face is recognised as well, it signifies that no face mask was used. However, if eyes are recognised but the face is not, it signifies that a face mask was worn by the person in consideration.
\par
Bu \emph{et al.} \cite{mref12} built a CNN-based cascaded face detector framework, consisting of three convolutional neural networks. The first CNN, Mask-1 is a very shallow fully convolutional layer network with 5 layers that gives a probability of being a masked face for each detection window, followed by a Non-maximum Supression (NMS) to merge overlapping candidates. Mask-2 is a deeper CNN with 7 layers, which resizes the candidate windows and also sets a detection threshold from the previous CNN. Mask-3 is also a 7 layer CNN which resizes the input windows it receives and gives a likelihood of whether it belongs to a masked face based on a preset threshold. After NMS, the remaining detection windows are the predicted detection results.
\par
Coming to more recent methodologies, Jiang \textit{et. al.} \cite{mref13} developed RetinaFaceMask, which is a novel framework for accurately and efficiently detecting face masks. The proposed framework is a one-stage detector which consists of a feature pyramid network to combine high-level semantic data with numerous feature maps. The authors propose a novel context attention module for the detection of face masks in addition to a cross-class object removal algorithm that discards predictions with low confidence values. The authors state that their model performs 2.3\% and 1.5\% more than the baseline result in face and mask detection precision respectively, and 11.0\% and 5.9\% higher than baseline for recall. \par

\section{Proposed Approach}
\begin{figure}
    \centering
    \includegraphics[width=50mm, height=100mm]{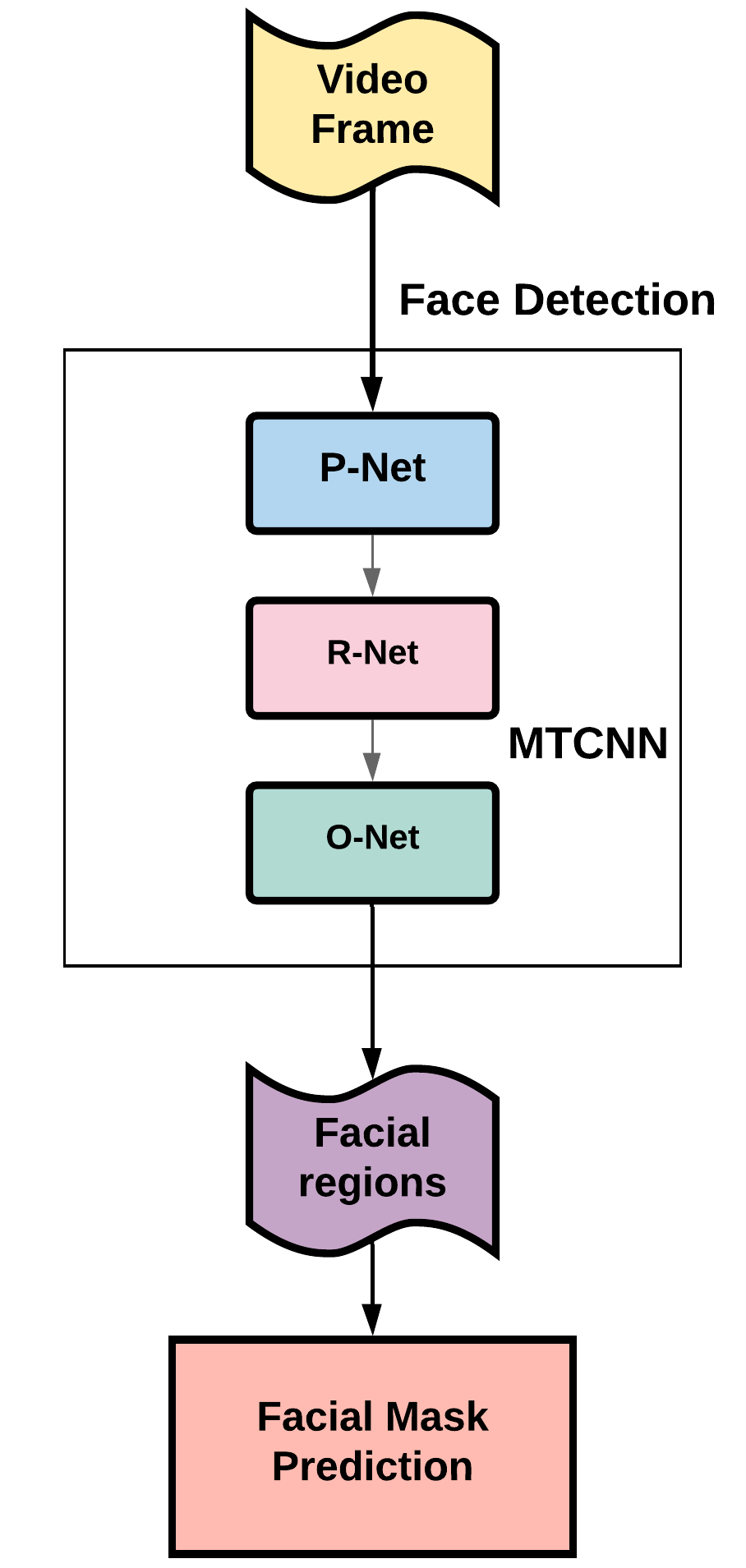}
    \caption{Workflow of proposed framework}
    \label{fig:propmethod}
\end{figure}

In this section, we elucidate our proposed framework, which is illustrated in Figure \ref{fig:propmethod}. The proposed framework aims to detect whether people in the video footage of a public area are wearing face masks or not. In order to do so, we first detect the face of the person and then determine if a facial mask is present on the face. It is to be noted that the terms `face mask' and `facial mask' are used interchangeably throughout this work.
\hfill \newline
\subsection{Face Detection}
\label{face_det}
For the task of face detection, we utilized the Multi-Task Cascaded Convolutional Neural Network (MTCNN)\cite{mref15} as the baseline model. The model is a cascaded structure comprising of three stages of deep convolutional networks that predict the facial landmarks.\par
The input image is initially resized to different scales in order to build an image pyramid, which behaves as input to the three-staged network elucidated below:
\begin{itemize}
\item \textbf{Stage 1} consists of a Fully Convolutional Network (FCN) called \textit{Proposal Network (P-Net)} \cite{mref15}, which is used to obtain the potential candidate windows in the input image pyramid and their bounding box regression vectors. In other words, \textit{P-Net} is responsible for proposing candidate facial regions from the input image. These estimated bounding box regression vectors are used to calibrate the candidate windows obtained, after which non-maximum suppression (NMS) is used to combine largely overlapping candidates.

\item \textbf{Stage 2} consists of a CNN called \textit{Refine Network (R-Net)} \cite{mref15} to which all the candidate windows obtained from the previous stage are fed. \textit{R-Net} mainly works to filter these candidate windows. This network rejects a large number of false candidates and utilises bounding box regression to calibrate the candidates obtained. For each candidate window, the offset between itself and the nearest ground-truth is predicted, denoted by $L_{i}^{box}$. The learning task is a regression problem and Euclidean loss is applied for each sample $x_{i}$ as:
\begin{equation}
 L_{i}^{box} = \norm{\hat{y}_{i}^{box} - y_{i}^{box}}_{2}^{2} 
\end{equation}
\par
where $\hat{y}_{i}^{box}$ is the target of the network and $y_{i}^{box}$ is the ground-truth coordinate.
\item \textbf{Stage 3} comprises of a CNN called \textit{O-Net} \cite{mref15}, which is responsible for proposing facial landmarks from the candidate facial regions obtained from the previous stage. \textit{O-Net} outputs facial landmark locations, namely the eyes, nose, and mouth regions of the face. Similar to the task of bounding box regression, the detection of facial landmarks is a regression problem and the following Euclidean loss is minimised:
\begin{equation}
  L_{i}^{landmark} = \norm{\hat{y}_{i}^{landmark} - y_{i}^{landmark}}_{2}^{2}  
\end{equation}
where $\hat{y}_{i}^{landmark}$ is the facial landmark coordinate predicted by the network and $y_{i}^{landmark}$ is the ground-truth coordinate.
\end{itemize}
\par 
For the task of face classification, the learning target can be formulated as a binary classification
problem. \par For each sample $x_{i}$, cross-entropy loss used was:
\begin{equation}
    L_{i}^{det} = -(y_{i}^{det}\log p_{i} + (1 - y_{i}^{det})(1 - \log p_{i}) )
\end{equation}
where $p_{i}$ is the probability produced by the network that the sample was a face and $y_{i}^{det} \in \{0, 1\}$ is the ground-truth label. \par
The output of this stage is the spatial coordinates of the bounding boxes enclosing the facial regions of the subjects in the frame.\par

\subsection{Facial Mask Prediction}
\label{mask_class}
For the task of identifying faces which are covered by a facial mask, we utilised the MobileNetV2 architecture \cite{mref14},  which is an effective feature extractor for object detection and segmentation. MobileNetV2 was chosen due to its ability to be deployed effortlessly on edge devices.\par
\begin{figure}[h]
    \centering
    \includegraphics[width=45mm, height=90mm]{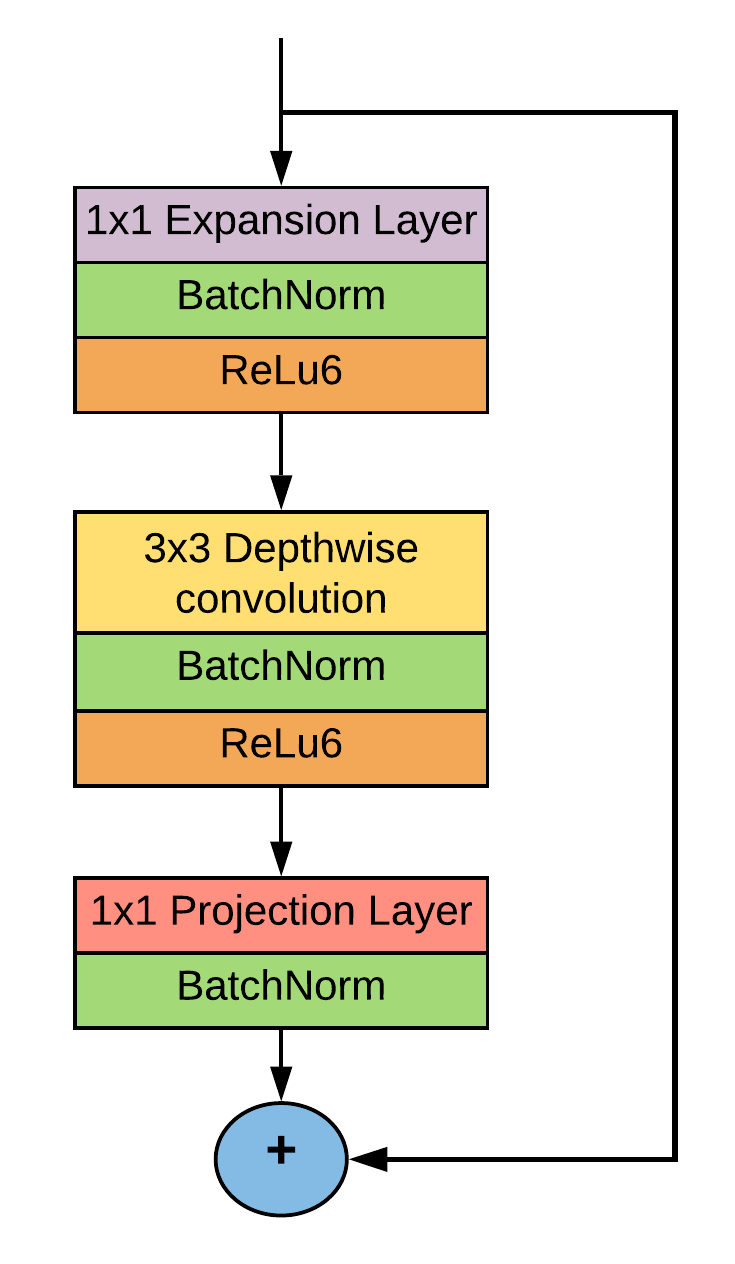}
    \caption{Bottleneck Residual block}
    \label{fig:block}
\end{figure}

MobileNetV2 uses depth-wise separable convolutions much like its predecessor, but the main residual block has some key alterations from its predecessor \cite{mref16}. The new residual block in MobileNetV2, known as the bottleneck residual block is illustrated in Figure \ref{fig:block}. There are a total of 3 convolutional layers in a block, where the latter two are: a depth-wise convolution that filters the input and a 1$\times$1 point-wise convolution. However, this 1$\times$1 convolution is quite different. This projection layer projects input data with a higher number of dimensions (channels) into a tensor with a much lower number of dimensions. As this layer suppresses the amount of data that flows through the network and the output of each block is a bottleneck, it is known as a bottleneck residual block. Hence, the input and output of the block are low-dimensional tensors whereas the filtering that takes place inside the block is on high-dimensional tensors. The other key aspect of MobileNetV2 is the residual connection. This primarily aids with the flow of gradients through the network during backpropagation.\par
Each layer has batch normalisation and the activation function used is ReLU6. However, an activation function is not applied to the output of the projection layer. Since this layer outputs low-dimensional data, succeeding this layer with non-linearity could destroy valuable information.\par

\begin{figure}[h]
    \centering
    \includegraphics[width=70mm, height=120mm]{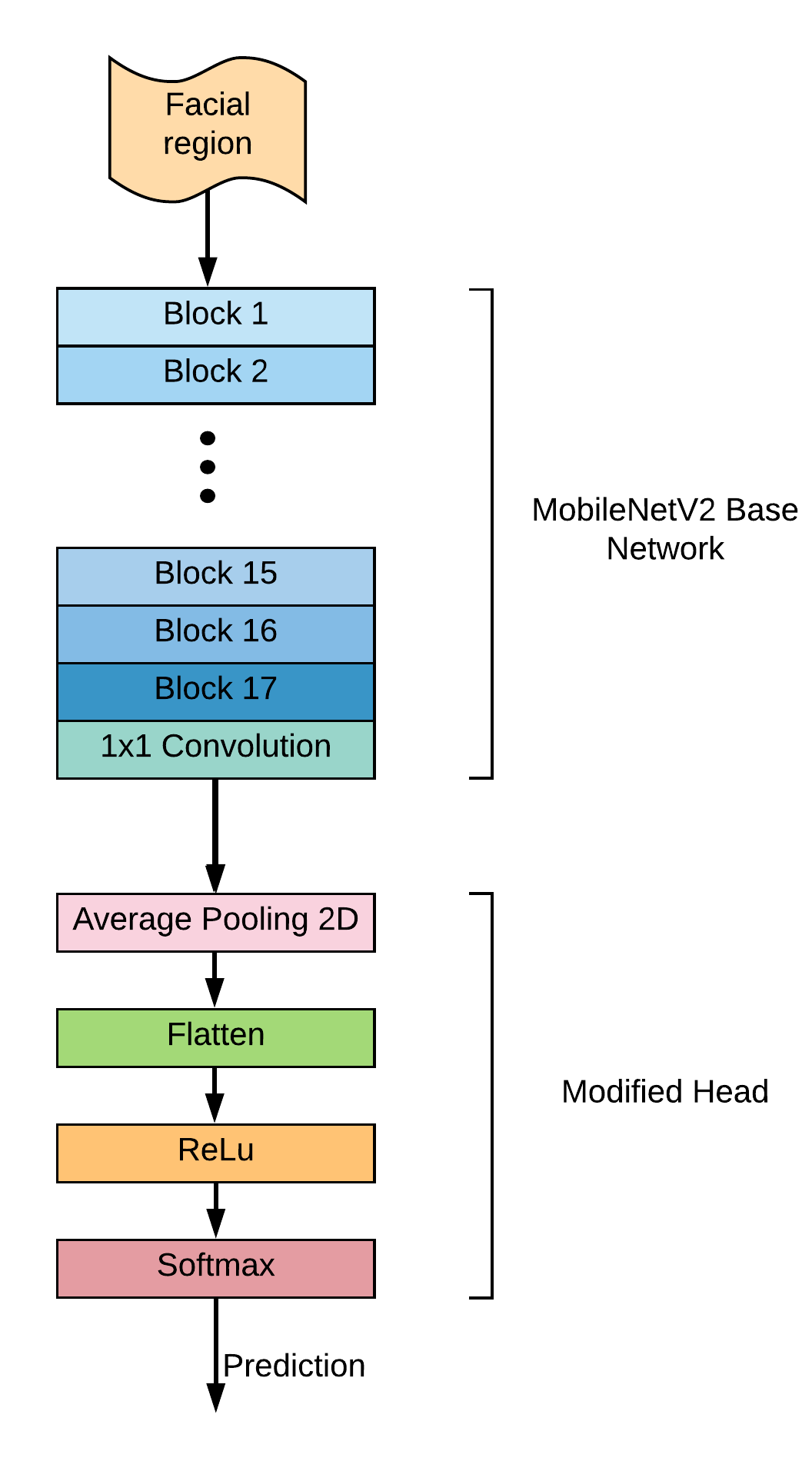}
    \caption{Facial mask classifier constructed using MobileNetV2 architecture}
    \label{fig:mobilenet}
\end{figure}

The full MobileNetV2 architecture, as illustrated in Figure \ref{fig:mobilenet}, comprises of 17 bottleneck residual blocks in a row. This is followed by a regular 1$\times$1 convolution. We utilise this base model of the MobileNetV2 architecture as a feature extractor for facial mask detection. We create a \textit{facial mask classifier} using 4 layers, succeeding the earlier mentioned architecture. We downsample each 2$\times$2 feature map using the average pooling layer (i.e. they are flattened) to produce a single long feature vector for classification. After passing through a ReLU activation function, we use a softmax function as illustrated in \ref{fig:mobilenet} to get the probability distribution over the predicted classifications. This is how the \textit{facial mask classifier} is able to predict whether a subject in a given frame is wearing a facial mask or not.\par

The facial regions obtained from the face detection model discussed in Face Detection (Section \ref{face_det}) are passed as input to the aforementioned \textit{facial mask classifier} and the output is a bounding box over each face region, with the label `Mask' indicating the presence of a face mask or `No Mask' when no face mask is worn by the subject in consideration. This output is illustrated in Figure \ref{fig:results}.
\begin{figure}
    \centering
    \includegraphics[width = 80mm, height = 45mm]{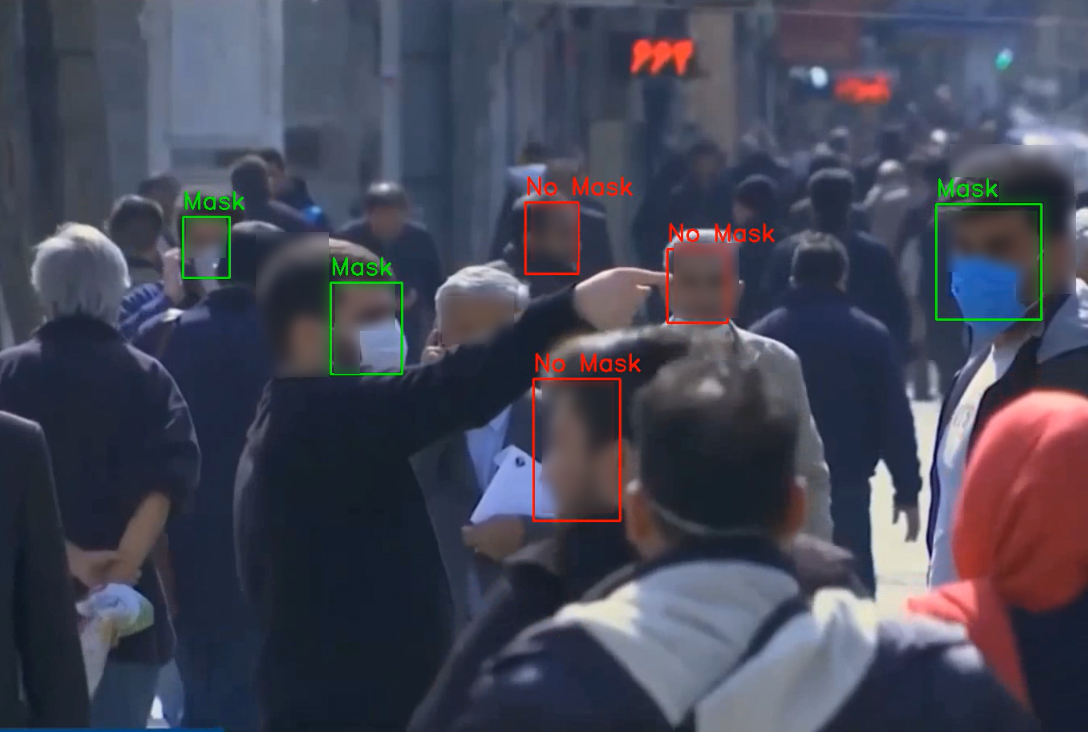}
    \includegraphics[width = 80mm, height = 45mm]{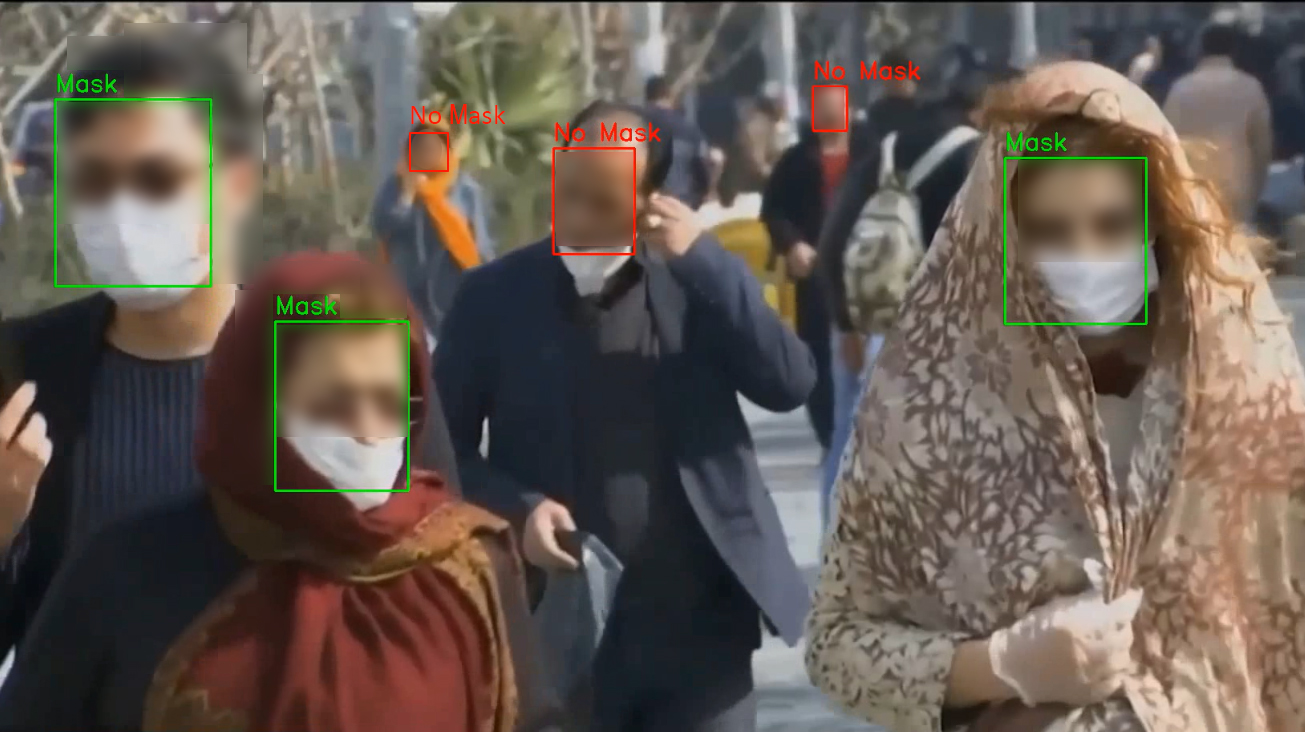}
    \caption{Visualisation of the results obtained by the proposed approach}
    \label{fig:results}
\end{figure}

\section{Experimental Evaluation}
In this section, we discuss the dataset used for conducting this study and the results obtained by the proposed approach. The experiments were conducted on Google Colab \cite{refx1} with Intel(R) Xeon(R) 2.00 GHz CPU, NVIDIA Tesla T4 GPU, 16 GB GDDR6 VRAM and 13 GB RAM. All programs were written in \textit{Python} - 3.6 and utilised \textit{OpenCV} - 4.2.0, \textit{Keras} - 2.3.0 and \textit{TensorFlow} - 2.2.0.\par

\subsection{Dataset Used}
\begin{figure}[!ht]
    \centering
    \includegraphics[width=80mm, height=80mm]{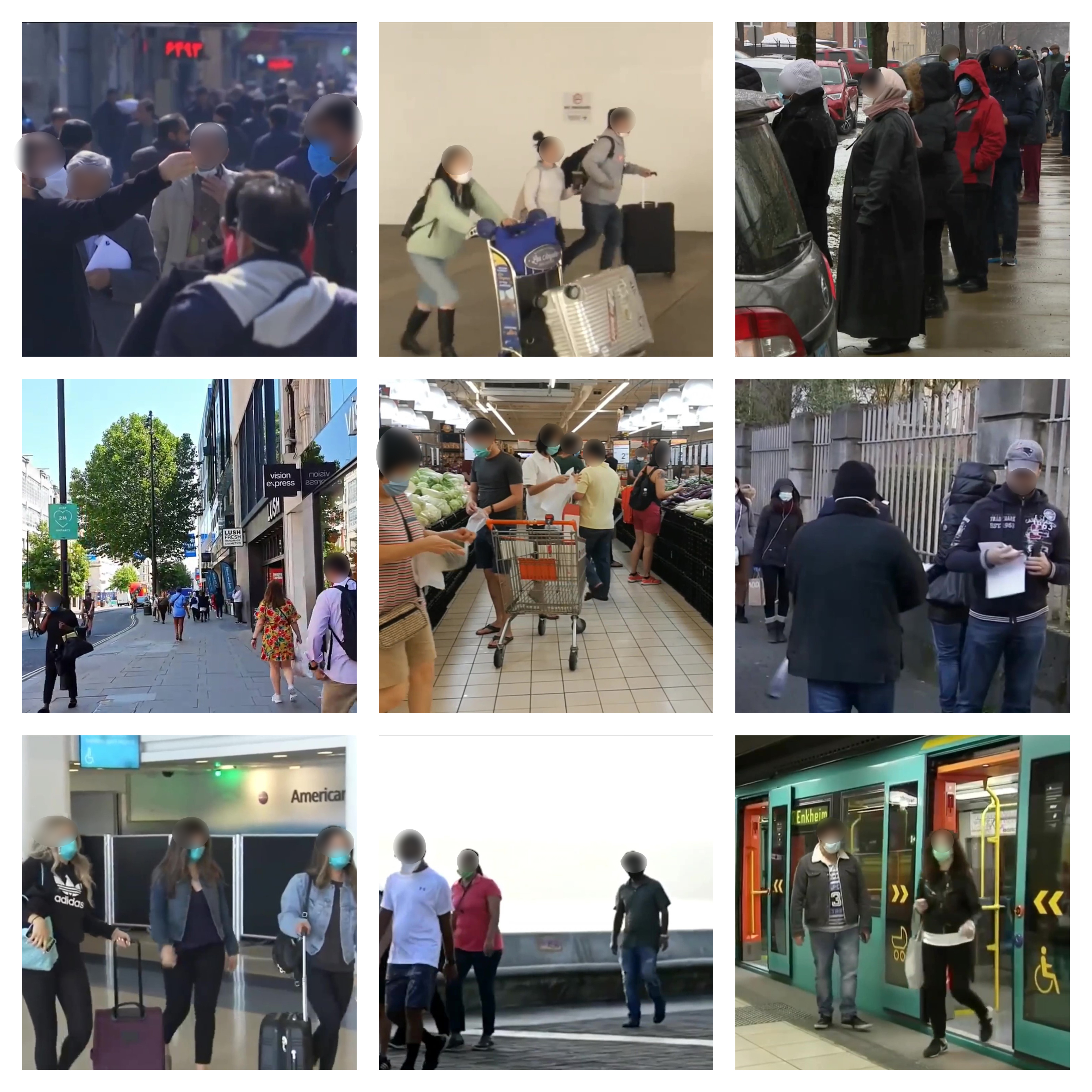}
    \caption{Some samples from the video dataset used in this work}
    \label{fig:dataset}
\end{figure}

The dataset used in this work is a collection of footage videos of public places from multiple geographical locations, compiled from YouTube. There are a total of 15 video samples in the dataset, each with an average duration of 1 minute. The videos capture the movement of people in public areas after the imposition of various safety rules and regulations in wake of the COVID-19 pandemic. The videos showcase people from multiple ethnicities and also capture different types of face masks worn by the public. Our dataset contains videos captured using different specifications of cameras and has a multitude of camera angles, varying illumination conditions, noise, and an average frames per second (FPS) of 30. Figure \ref{fig:dataset} illustrates a few sample videos present in this dataset.\par

\begin{figure*}[!ht]
\centering
\begin{minipage}{1\textwidth}
    \centering
    \includegraphics[width=0.99\linewidth, height=0.42\textheight]{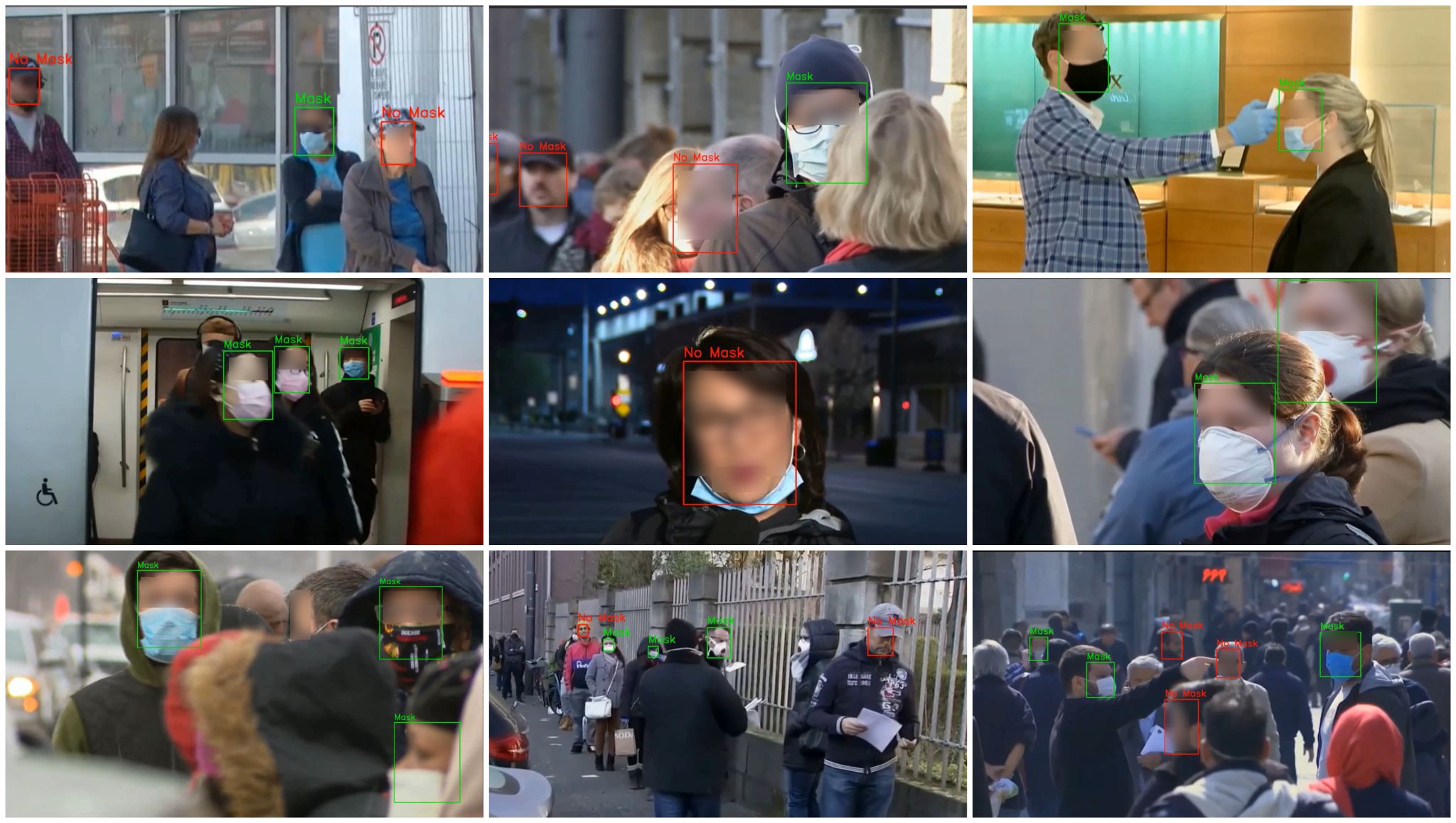}
    \caption{Some instances of the results obtained by the proposed approach}
    \label{fig:results}
    \end{minipage}%
\end{figure*}

\subsection{Experimental Results and Statistics}
The proposed approach has been evaluated by measuring the precision, recall, and accuracy metrics of the face detection model and \textit{facial mask classifier} respectively.\par

\begin{gather}
    \textit{Precision} = \frac{TP}{TP+FP} \times 100\% \label{eq:prec}\\   
    \textit{Recall} = \frac{TP}{TP+FN} \times 100\% \label{eq:recall}\\
    \textit{Accuracy} = \frac{TP+TN}{TP+TN+FP+FN}\times 100\% \label{eq:metrics}
\end{gather}

where \textit{TP, TN, FP and FN} denote the true positive, true negative, false positive, and false negative observations respectively.
\subsubsection{Face Detection}
The face detection model mentioned in Section \ref{face_det} achieved a \textit{precision} of 94.50\%, \textit{recall} of 86.38\%, and \textit{accuracy} of 81.84\% on the chosen dataset. \par

\subsubsection{Facial Mask Prediction}
The \textit{facial mask classifier} mentioned in Section \ref{mask_class} achieved a \textit{precision} of 84.39\%, \textit{recall} of 80.92\%, and \textit{accuracy} of 81.74\% on the chosen dataset.\par

\begin{table}[h]
\centering
\caption{Comparison of proposed framework with Cascaded framework for mask detection \cite{mref12}}
\label{tab:table1}
\begin{tabular}{|l|l|l|}
\hline
Approach                                                                         & Accuracy & Recall \\ \hline
Proposed Framework                                                               & 81.74\%         & 80.92\%       \\ \hline
\begin{tabular}[c]{@{}l@{}}Cascaded framework\\  for mask detection\end{tabular} & 86.6\%          & 87.8\%       \\ \hline
\end{tabular}
\end{table}

\begin{table}[h]
\centering
\caption{Comparison of proposed framework with RetinaFaceMask \cite{mref13}}
\label{tab:my-table}
\begin{tabular}{cclcl}
\hline
\multicolumn{1}{|c|}{\multirow{2}{*}{Approach}} & \multicolumn{2}{c|}{Face}               & \multicolumn{2}{c|}{Mask}               \\
\multicolumn{1}{|c|}{}                          & Precision& \multicolumn{1}{l|}{Recall} & Precision& \multicolumn{1}{l|}{Recall} \\ \hline
\multicolumn{1}{|c|}{Proposed Framework} & \multicolumn{1}{c|}{94.50\%} & \multicolumn{1}{l|}{86.38\%} & \multicolumn{1}{c|}{84.39\%} & \multicolumn{1}{l|}{80.92\%} \\ \hline
\multicolumn{1}{|c|}{\begin{tabular}[c]{@{}l@{}}RetinaFaceMask\\ with MobileNet\end{tabular}}       & \multicolumn{1}{c|}{83.0\%} & \multicolumn{1}{l|}{95.6\%} & \multicolumn{1}{c|}{82.3\%} & \multicolumn{1}{l|}{89.1\%} \\ \hline
                                                &           &                             &           &                            
\end{tabular}
\end{table}

Table \ref{tab:table1} compares our proposed framework to the cascaded framework used in \cite{mref12}. The higher accuracy of the cascaded framework is due to the fact that it was designed to work on images rather than videos. Also, the ``MASKED FACE'' dataset \cite{mref12} which was used to test the cascaded framework comprises of people wearing head gear. On the other hand, the dataset used to evaluate our proposed framework captures the various types of face masks worn by the public as a precautionary measure for disease control.\par
Table \ref{tab:my-table} compares our proposed framework to RetinaMask \cite{mref13}. It can be observed that our proposed framework achieves a higher precision value in detecting masks and faces as compared to RetinaMask. However, RetinaMask achieves a higher recall as the dataset it was evaluated on comprises of images of a close-up of people's faces which accounts for their better recall figures in detecting masks and faces. Also, the authors of RetinaMask do not mention the effectiveness of their model in detecting multiple faces at once, while our model works well in detecting multiple faces, as illustrated in Figure \ref{fig:results}.\par
Finally, our proposed framework has also been tested on a video dataset unlike the aforesaid approaches which deal with image datasets. The video dataset used to evaluate the proposed framework contains videos taken using different specifications of cameras and has a multitude of camera angles, varying illumination conditions and noise. Thus, the proposed approach will perform well on real world camera captures.\par

\subsection{Analysis of the proposed approach}
From the earlier discussion, it can be observed that the effectiveness of the \textit{facial mask classifier} depends on the effectiveness of the face detection model. If the face detection model fails to detect a face or incorrectly identifies an object as a face, the performance of the \textit{facial mask classifier} is affected.\par
The following key observations were made about the effectiveness of the proposed approach:
\begin{enumerate}
    \item It is able to detect facial masks on subjects present at a considerable distance from the camera.
    \item It performed well even in scenarios where the public areas captured were crowded.
    \item It satisfactorily detected the presence of facial masks on subjects not directly facing the camera (i.e. only a side profile of the face was visible) in most cases.
    \item It was able to identify subjects who were incorrectly wearing a facial mask (i.e. the mask was not covering their mouth and nose) and labeled them as `No Mask'.
   
\end{enumerate}
These observations are illustrated in Figure \ref{fig:results}.
\section{Conclusions and Future Work}
In this work, a new approach for detecting face masks from videos is proposed. A highly effective face detection model is used for obtaining facial images and cues. A distinct facial classifier is built using deep learning for the task of determining the presence of a face mask in the facial images detected. The resulting approach is robust and is evaluated on a custom dataset obtained for this work. The proposed approach was found to be effective as it portrayed high \textit{precision}, \textit{recall}, and \textit{accuracy} values on the chosen dataset which contained videos with varying occlusions and facial angles. The effectiveness of the facial mask classifier largely confides on the ability of the face detection algorithm to accurately identify faces in the video frames. This could be the subject of future research in this direction.

\section{Acknowledgement}
We thank Google Colaboratory for providing access to computational resources used for this study and YouTube for helping us avail the videos used in our dataset. We also thank our institute, the National Institute of Technology Warangal for its constant support and encouragement to undertake research.

\bibliographystyle{IEEEtran}
\bibliography{main}
\end{document}